\pdfoutput=1

\documentclass[11pt]{article}

\usepackage[final]{acl}

\usepackage{times}
\usepackage{latexsym}

\usepackage[T1]{fontenc}
\usepackage[utf8]{inputenc}

\usepackage{microtype}

\usepackage{inconsolata}

\usepackage{graphicx}

\usepackage{lipsum}
\usepackage{booktabs}
\usepackage{bm}
\usepackage{amsmath}
\usepackage{mathtools}
\usepackage{multirow}
\usepackage{adjustbox}
\usepackage{lengthconvert}
\usepackage{pifont}
\usepackage{tabularx}
\usepackage{array}
\usepackage{csquotes}
\usepackage{subcaption}
\usepackage{xparse} %
\usepackage{pgffor} %
\usepackage{mfirstuc} %
\newcommand{\xmark}{\ding{55}}%

\newcommand{\argmax}{\mathop{\mbox{argmax}}}
\newcolumntype{L}{>{\small}l}

\ExplSyntaxOn
\tl_new:N \l_tmpamacro_tl
\NewDocumentCommand{\lastwords}{m m}
 {
  \seq_set_split:Nnn \l_tmpa_seq { ~ } { #2 }
  \int_compare:nTF { \seq_count:N \l_tmpa_seq > #1 }
   {
    \ldots\
    \int_step_inline:nn { \seq_count:N \l_tmpa_seq - #1 }
     {
      \seq_pop_left:NN \l_tmpa_seq \l_tmpamacro_tl
     }
    \seq_use:Nn \l_tmpa_seq { ~ }
   }
   {
    #2
   }
 }
\ExplSyntaxOff

\title{MLLP-VRAIN UPV system for the IWSLT 2025 Simultaneous Speech Translation Translation task}

\newcommand{\upvdir}{Machine Learning and Language Processing, VRAIN, Universitat Politècnica de València}
\newcommand{\uvdir}{Departament d’Informàtica, Escola Tècnica Superior d’Enginyeria, Universitat de València}
\author{
    Jorge Iranzo-Sánchez*$^{\diamond}$ \and Javier Iranzo-Sánchez$^{\diamond}$ \\
    \and {\bf Adrià Giménez$^\dagger$} \and {\bf Jorge Civera$^\diamond$} \and {\bf Alfons Juan$^\diamond$}
   \\
   $^\diamond$\upvdir\\
   $^\dagger$\uvdir
   \\
   \texttt{\{jorirsan,jairsan,jorcisai,ajuanci\}@upv.es},\space \texttt{adria.gimenez@uv.es}
 }

\begin{document}
\maketitle
\begin{abstract}
This work describes the participation of the MLLP-VRAIN research group in the shared task of the IWSLT 2025 Simultaneous Speech Translation track. Our submission addresses the unique challenges of real-time translation of long-form speech by developing a modular cascade system that adapts strong pre-trained models to streaming scenarios. We combine Whisper Large-V3-Turbo for ASR with the multilingual NLLB-3.3B model for MT, implementing lightweight adaptation techniques rather than training new end-to-end models from scratch. Our approach employs document-level adaptation with prefix training to enhance the MT model's ability to handle incomplete inputs, while incorporating adaptive emission policies including a wait-$k$ strategy and RALCP for managing the translation stream. Specialized buffer management techniques and segmentation strategies ensure coherent translations across long audio sequences. Experimental results on the ACL60/60 dataset demonstrate that our system achieves a favorable balance between translation quality and latency, with a BLEU score of 31.96 and non-computational-aware StreamLAAL latency of 2.94 seconds. Our final model achieves a preliminary score on the official test set (IWSLT25Instruct) of 29.8 BLEU. Our work demonstrates that carefully adapted pre-trained components can create effective simultaneous translation systems for long-form content without requiring extensive in-domain parallel data or specialized end-to-end training.
\end{abstract}

\section{Introduction}\label{sec:intro}

\begin{figure}[ht]
  \centering
  \includegraphics[width=\columnwidth]{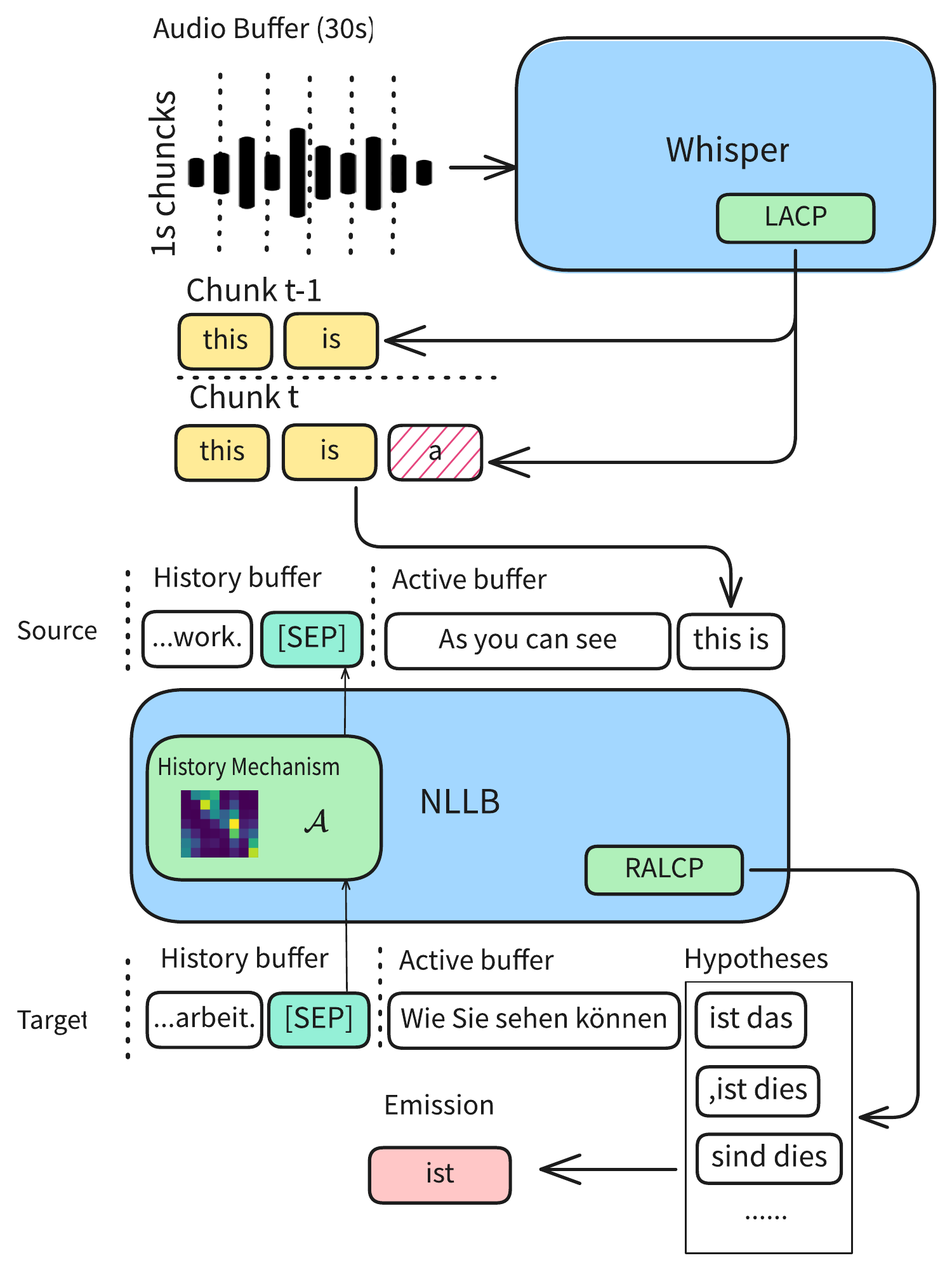}
  \caption{System diagram of our cascaded system for the SimulST track.\label{fig:system}}
\end{figure}

In this paper we describe the participation of the MLLP-VRAIN research group in the shared tasks of the 22th International Conference on Spoken Language Translation (IWSLT)~\citep{abdulmumin-etal-2025-findings}.
We participated on the Simultaneous Speech Translation (SimulST) task in the English to German direction. Compared to other years, two aspects were changed in the shared task which guided
the construction of our system: The usage of pretrained open weight models and the evaluation of long-form audio. Our participation this year was an attempt of creating a production ready model
based on the minimal adaptation of offline ASR and MT~\cite{papiDoesSimultaneousSpeech2022}. Recent years have seen a rise in the usage of end-to-end approaches\footnote{As defined by IWSLT~\url{https://iwslt.org/2025/offline\#evaluation-conditions}.} which, in theory, can offer a better integration and can avoid error compounding between ASR and MT components. However, they typically require large quantities of parallel speech-to-text translation data, which is often scarce and costly to obtain. Cascade systems, on the other hand, while they are more data-efficient due to the abundance of separate ASR and MT training resources, may suffer from error propagation and lack of joint optimization.
Nevertheless, recent shared tasks and evaluation campaigns continue to show that cascade systems generally achieve superior performance over current end-to-end alternatives~\cite{iwslt-2023-international, iwslt-1-2024}.
As such, we model our system based on the cascaded approach, in which we take a special keen interest due to its inherent modularity and easier reuse of strong pre-trained components. Figure~\ref{fig:system} shows the overall architecture of our system.

\section{System Architecture}
\subsection{ASR system}
For the choice of the ASR components, we select our model based on the results of public ASR systems on common benchmarks available at the Hugging Face Open ASR Leaderboard~\citep{open-asr-leaderboard}.
After some initial tests and taking into account our computing limitations, we selected Whisper-Large-V3-Turbo~\citep{RadfordKXBMS23}~\footnote{\url{https://huggingface.co/openai/whisper-large-v3-turbo}} as our final ASR model. We use the model as it is, and we do not make usage of any of the English audio data provided by the organizers for finetunning of the model. 
There are two reasons for this decision. First, when looking at the provided datasets, we conjecture that the Whisper model has probably have already seen this data; and second, that we fear that we may end up lowering the performance of the system as there is a domain mismatch between the provided datasets and evaluation set, with the latter being scientific talks of the ACL.

The adaptation for streaming is done in a similar way to the one described in \citet{machacek-etal-2023-turning}, where a Longest Common Prefix policy along some heuristics are combined for the usage in a streaming fashion.
Inference is done via the Faster-Whisper library~\footnote{\url{https://github.com/SYSTRAN/faster-whisper}}.
We select a maximum audio buffer of 30 seconds and a minimum chunk size of one second and deactivate the usage of VAD filtering. The audio buffer is cleaned when a end of sentence is detected by external sentence splitter
or the 30 seconds window is full. The audio stream is then updated accordingly to the timestamps obtained by the typical DTW procedure used in Whisper.

During development we detected that this base system had sometimes very unreliable behaviour due to latency spikes derived from indecisions of the LCP policy on the punctuation, casing and styling of some words. 
To alleviate this, we relaxed the LCP policy so that it is done on lower cased input, with no punctuation signs, and with the additional constraint of a threshold by Levenshtein Distance to consider if a word in the prefix is sufficiently different.
We select a Levenshtein Distance threshold of two so that if the distance of the words to be checked is less or equals to it, then they are to be considered the same by the policy. 

The ASR system obtains a WER of 8.54\% on the ACL60/60~\citep{salesky-etal-2023-evaluating} development set. 
\subsection{MT system}
During recent years a series of works have appeared exploring realistic scenarios for SimulST where models are evaluated on long-form speech scenarios. 
\citep{schneider-waibel-2020-towards, sen2022simultaneoustranslationunsegmentedinput, papi-etal-2024-streamatt,polakLongFormEndtoEndSpeech2023, iranzo-sanchez-etal-2024-segmentation, ouyang2025infinisstsimultaneoustranslationunbounded}.
Taking this into consideration, for our MT model, we adapt an offline MT model to streaming through a lightweight procedure inspired to that of~\citet{iranzo-sanchez-etal-2024-segmentation}, which presents an easy to adapt pipeline for the creation of an MT component in SimulST system which we further adapt and simplify it to our given conditions.
Overall, through training with prefix-training~\cite{NiehuesPHSW18} and document level metadata, the model is able to learn to work with an incomplete stream and control a history stream~\cite{iranzo-sanchez-etal-2022-simultaneous} by emitting a sentinel token [SEP] which then serves to call a lightweight log-linear model to take into account which part of the stream has already been translated. When a certain threshold of size in the buffer reached, the oldest pairs of identified segments can easily be discarded without any fear of possible mismatches in the stream. 

\paragraph{Model:} 
Instead of training from scratch, we make use of the multilingual system NLLB-3.3B~\citep{NLLB_Team24}. We also explored the usage of MADLAD-400~\citep{KuduguntaC0GXKS23}, but discarded it after founding BLEU scores of over 95 points when evaluation of the system in an offline scenario for the ACL60/60 dataset, which indicated a possible data contamination of the test set. We also tested some LLMs, but found early results too unsatisfactory for our computing budget. 

\paragraph{History Mechanism:} Inspired by the usage of attention maps of~\citet{papiAlignAttUsingAttentionbased2023}, for the log-feature model, instead of the described reverse translation model of the original paper, we found that a single feature which looks at the most attended position of the previous token before the sentinel [SEP] token
to be more simpler and effective to determine the segmentation position of the source stream. To be more precise, the position of the last source word $\hat{a}$ to be moved to the streaming history buffer for the current active source and target chunks $x$ and $\hat{y}$ is
    \begin{equation}
    \hat{a} = \argmax_i{\mathcal{A}}(x_{i},y_{[SEP]-1})
    \end{equation}
with $\mathcal{A}$ being the attention score function and $i$ indicates the position of source word $x$ in the active chunk.

\paragraph{Data and Training:} 
For the document level data, we take the available News Comentary\footnote{\url{https://data.statmt.org/news-commentary/v18/training/}} and Europarl\footnote{\url{https://www.statmt.org/europarl/v10/training/europarl-v10.de-en.tsv.gz}} datasets with their reconstructed document level information extracted from Paradocs~\citep{wicks-etal-2024-recovering} for a total of 36225 documents. While not in the target domain of ACL talks, we hope that our adaptation is able to move the model from offline inference domain to simultaneous with the usage of history context. Additionally, we want to adapt the multilingual model so that it can better focus on the English to German direction. 
Prefix augmented data with document level information is created as in the original work, but is dynamically generated at training time. For each document, which represents a data sample, we select a sentence and randomly prepend from 1 up to 10 of the previous phrases of the document with the corresponding sentinel token [SEP]. Then, for the corresponding phrase we create the prefix by taking into account the ratio of the length of the active source and target phrase. During training, the usage of prefix training is triggered at a rate of 50\%.
As for the training procedure, we make use DoRA~\citep{MaoHGBMX24} to fine-tune our MT model. Training hyperparameters are shown in Appendix~\ref{sec:appendix}.

\paragraph{Policy:} For our policy in the MT component, we make use of RALCP~\citep{wang-etal-2024-simultaneous} using the hypotheses of the system beam search in combination with a \emph{wait-k} policy~\citep{ma-etal-2019-stacl}, the latter being active only during the beginning of a new phrase (that is after emitting a [SEP] token) to prevent system hallucination. We also clean \enquote{invalid} empty beam hypotheses, which were very frequent on the baseline model, adjusting the RALCP $\lambda$ accordingly so that the ratio of hypotheses and $\lambda$ stays roughly the same as before filtering.

The final peak memory usage during inference of both ASR and MT models combined is of 20GB of VRAM~\footnote{On a Nvidia GTX 4090 with a Intel(R) Core(TM) i9-10920X CPU @3.50GHz}, with values fluctuating between 1-3GiB depending on the current audio and translation history buffers. We leave further optimization for future work.

\section{Evaluation}
For our baselines, we searched for any other public simultaneous speech translation systems capable of long form translation. However, we only found the system described in~\citet{papi-etal-2024-streamatt} available, 
and in this case, we observed that the system had strong hallucinations that ended up in an unrecoverable state for long enough audios. Due to this, we instead build a \enquote{naive} baseline for our system which make use of the plain translation system without any additional training. For controlling the history buffers, we simply remove from the source and target text history buffers after a maximum number of words is reached in each buffer. In practice, we surpassingly found that while the system will end up with slight mismatches between the source and target stream at the head position, systems are still able to work in a streaming scenario. We additionally add the offline inference of the model before and after fine-tuning and the best baseline from the IWSLT organizers that achieved a similar latency-quality tradeoff matching our own.\footnote{Organizer baseline results extracted from \url{https://github.com/pe-trik/iwslt25-baselines/tree/master/experiments/acl6060\_dev/de/cascade}}. Best hyperparameters with the optimal quality/latency are shown in Appendix~\ref{sec:appendix}.

Table~\ref{tab:results} shows the results of the baseline and our adapted model,
the stream LAAL, both computationally an computationally aware,
BLEU~\cite{papineni-etal-2002-bleu} as calculated with SacreBLEU~\cite{post-2018-call}\footnote{Python3.12.9|BLEU|nrefs:1|case:mixed|eff:no|tok:13a|\\smooth:exp|version:2.5.1} and the COMET-22~\citep{rei-etal-2022-comet}\footnote{Python3.12.9|Comet2.2.6|fp16|Unbabel/wmt22-comet-da|r1}. We follow the recomendations of \citet{zouhar-etal-2024-pitfalls} and set the COMET score to 0 for samples where the target translation is empty after re-segmentation with mWERSegmenter~\cite{matusov-etal-2005-evaluating}.

\begin{table}[t]
\begin{center}
    \begin{adjustbox}{max width=\columnwidth}
    \begin{tabular}{l c    c     c           c}
        \toprule
        &   &   & \multicolumn{2}{c}{StreamLAAL}\\
    \toprule
                 & BLEU & COMET & NCA & CA \\ \midrule
       Offline & 43.12 & 0.833 & --- & --- \\
       Offline$^{(A)}$ & 41.48 & 0.836  & --- & --- \\ \midrule
       Baseline$^{iwslt}$ & 25.47 &  ---  & 3.67    & --- \\
       Baseline$^{upv}$ & 26.10 &  0.642  & 3.61    & 4.35 \\
       Adapted & 31.96 &  0.732 &  2.94 &   4.20 \\
    \bottomrule
    \end{tabular}
     \end{adjustbox}
    \caption{Quality (BLEU, COMET$\uparrow$) and non-computational and computational aware (NCA/CA) latency (StreamLAAL (secs)$\downarrow$) results on the ACL60/60 development set. Offline models take the golden reference source text.
    The Offline$^{(A)}$ model refers to the results of our adapted model when doing offline inference. Offline models results are obtained given the golden source reference transcription.~\label{tab:results}}
\end{center}
\end{table}

First of all, impact of the translation quality degradation in the offline mode seems to minimal, with deltas of 1.64 BLEU and 0.03 of COMET respectively.
As for our baseline and adapted models, we can see that there is a significant quality degradation. We can attribute this to multiple factors, but we would like to highlight two aspects. First, is that we are comparing ourselves to offline systems that take the golden reference transcription, and thus, transcriptions errors from the ASR and possible resegmentation errors introduced by StreamLAAL are not taken into account in the evaluation of our offline baseline. The second factor which may explain this gap is the mere nature of the ACL60/60 dataset. We would like to highlight this one in particular since the translations where originally based on post-edits from offline translation models and thus more suited for the evaluation of offline speech translation compared to that of SimulST. Thus, we think that the resulting translations from our SimulST system, which should be more monotonic in nature compared to the offline baselines and ACL60/60 references, maybe be more penalized in a similar way to the observations of~\citet{doi-etal-2024-word}.

In terms of baselines, we can see that our naive baseline slight beats the organizers baseline in the selected quality-latency range. When comparing our baseline and adapted model, we can see a considerable increase in both quality and reduction of latency.
Our adapted model ends up scoring 31.96 BLEU and 0.732 COMET and StreamLAAL scores of 2.94 and 4.20 seconds depending on the computational awareness of the metric calculation. This places our model on the high latency regime as defined by the shared task description. 

To better study the latency of our system, Table~\ref{tab:percent} shows the top percentiles of StreamLAAL as well as their medians and maximum recorded values. We can see how for all of these metrics our adapted system consistently beats our baseline and ensures a better performance on the worst case scenarios, with these delays being the more impactful for the end user of SimulST systems.

An observation that can be made in this table is that of the considerable increase of latency for the worst cases between the NCA and CA metrics.
After investigating, we discovered the temperature fallback mechanism of Whisper to seem to cause this phenomena, resulting in some rare cases where latency spikes occur and can only be observed when taking computational costs into account.
Despite this, we found that in practice the performance is really poor with this feature disabled. In general, we observed that the ASR emission policy highly influenced the system performance, with hyperparameter changes on the MT system having less of an overall impact.

Regarding the official test set (IWSLT25Instruct), preliminary results by the organizers indicate that our model achieved a final score of 29.8 BLEU.

\begin{table}[t]
\centering
\begin{tabular}{@{}l@{~}r@{~~~}r@{~~~}r@{~~~}r@{~~~}r@{~~~}r@{}}
\toprule
 model &   M      & mdn & p90 & p95 & p99 & max \\
\midrule
& \multicolumn{6}{c}{NCA}\\
\cmidrule{2-7}
 Baseline & 3.61 & 2.96  &  6.57 & 9.14  &  13.51 &  16.59 \\
 Adapted  & 2.94 & 2.65  &  4.30 & 5.38  &  7.59  &  9.25\\
\cmidrule{2-7}
& \multicolumn{6}{c}{CA}\\
\cmidrule{2-7}
 Baseline & 4.35 &  3.62  &  7.65  &  10.51 &  14.58 &  18.52\\
 Adapted  & 4.20  & 3.55  &  5.98  &  7.64 &   10.73 &  15.00 \\
\bottomrule
\end{tabular}
\caption{StreamLAAL mean (M), median (mdn), percentiles
90\%, 95\% and 99\%, and maximum value (in seconds) for the Baseline$^{upv}$ and Adapted system.\label{tab:percent}}
\end{table}

\section{Conclusions}
In this paper we described our SimulST system for the IWSLT 2025 Simultaneous Speech Translation task.
Preliminary results show that our cascaded based system using Whisper and NLLB showed a good performance
and achieved a good balance between translation quality and latency. We see how through adaptive policies
and very computationally cheap adaptation a long-form speech SimulST system can be created from offline models. 
Future work could be expand to see the robustness of this methodology, such as the usage of synthetic document level bitext data~\cite{post2024escapingsentencelevelparadigmmachine}
or speech data. Investigating more robust adaptive latency policies or techniques which
better optimize ASR and MT components~\citep{tran-etal-2022-joint} while preserving the benefits of the cascaded approach could further greatly enhance the system performance. 
Also, a gap still exists compared to offline translation, which should be further explore d more detail.

Additionally, the usage of LLMs to serve as all in one transcriber, translator and re-scorer in a cascaded pipeline along their robustness and usage of long context shows promising results for their usage in long-form speech translation if computational costs can be taken into account.

\section{Limitations}

Due to time constraints, hyperparameters search for trade-offs between translation and latency of models was limited, as well as the tuning of the ASR system.
In our participation, we restricted ourself to the English to German direction,
but we think that our approach could be generalized to the other language pairs in the competition.
We hope to participate in future editions covering all language pairs available and expanding the breadth and scale of the 
studied models.

\section*{Acknowledgments}
The research leading to these results has received funding from EU4Health
Programme 2021--2027 as part of Europe's Beating Cancer Plan under Grant
Agreements nos. 101056995 and 101129375; and from the Government of
Spain's grant PID2021-122443OB-I00 funded by
MICIU/AEI/\allowbreak10.13039/\allowbreak5011\-00011033 and by
``ERDF/EU'', and grant PDC2022-133049-I00 funded by
MICIU/AEI/\allowbreak10.13039/\allowbreak501100011033 and by the
``European Union NextGenerationEU/PRTR''. The authors gratefully
acknowledge the financial support of Generalitat Valenciana under project
IDIFEDER/\allowbreak2021/\allowbreak059.

\bibliography{anthology_0,anthology_1,anthology_2,anthology_3,anthology_4,anthology_5, custom, extra}

\appendix
\section{Hyperparameters}~\label{sec:appendix}

\begin{table}[ht]
\begin{center}
    \begin{adjustbox}{max width=\columnwidth}
    \begin{tabular}{l c c }
        \toprule
        Hyperparameter & Baseline & Adapted \\ \midrule
        ASR VAD & \multicolumn{2}{c}{\xmark} \\
        ASR Initial Wait & \multicolumn{2}{c}{1s} \\
        ASR LCP Chunk & \multicolumn{2}{c}{1s} \\
        ASR Beam Size & \multicolumn{2}{c}{5} \\
        MT Wait-k & \multicolumn{2}{c}{3} \\
        MT RALCP $\lambda$ & \multicolumn{2}{c}{0.5} \\
        MT Beam Size & \multicolumn{2}{c}{10} \\
        MT Attention Head Layer & \multicolumn{2}{c}{6}  \\
        MT Max Buffer & \multicolumn{2}{c}{80 words} \\
        MT History Remove & 20 words & 1 sentence \\
    \bottomrule
    \end{tabular}
     \end{adjustbox}
    \caption{Inference hyperparameters for the baseline and adapted models\label{tab:hyperparameters}}
\end{center}
\end{table}

\begin{table}[ht]
    \footnotesize
    \addtolength{\tabcolsep}{-1pt}
    \centering
    \begin{adjustbox}{max width=\columnwidth}
    \begin{tabular}{lc}
        \hline
        \toprule
        Hyperparameter  & Value      \\
        \midrule
			Optimizer    & 8bit-AdamW \cite{DettmersLSZ22} \\
                               Warm up Ratio & 0.06 \\
                               LR Schedule  & Linear \\
                               Effective Batch Size & 64   \\
        \midrule
                               Epochs &  3 or until convergence   \\
                               Initial Learning Rate & 2e-4 \\
                               DoRA Dropout & 0  \\ 
                               Target Modules & $\bm{Q, V}$ and Vocabulary Embeddings $\bm{E}$ \\
                               DoRA rank config. &  $r_{\bm{Q}}=r_{\bm{K}}=r_{\bm{E}}=16$  \\
                               DoRA $\alpha$ &  32 \\
                               Bias &  \xmark \\
        \bottomrule
    \end{tabular}
    \end{adjustbox}
    \caption{DoRA hyperparameters for the trained adapted model.\label{tab:hiper_lora}}
\end{table}

\end{document}